\documentclass{article}

\usepackage{PRIMEarxiv}

\usepackage[utf8]{inputenc} 
\usepackage[T1]{fontenc}    
\usepackage{hyperref}       
\usepackage{url}            
\usepackage{booktabs}       
\usepackage{array}
\usepackage{tabularx}
\usepackage{amsfonts}       
\usepackage{nicefrac}       
\usepackage{microtype}      
\usepackage{lipsum}
\usepackage{fancyhdr}       
\usepackage{graphicx}       
\usepackage{caption}        
\usepackage{xcolor}         
\usepackage{algorithm}      
\usepackage{algorithmic}   
\graphicspath{{media/}}     

\title{From Topology to Trajectory: LLM-Driven World Models for Supply Chain Resilience
\thanks{\textit{\underline{Citation}}: 
\textbf{Authors. Title. Pages.... DOI:000000/11111.}} 
}

\author{
  Jia Luo \\
  Huazhong University and Science and Technology \\
  Wuhan, China\\
  \texttt{u202317016@hust.edu.cn} \\
}

\begin{document}
\maketitle

\begin{abstract}
Semiconductor supply chains face unprecedented resilience challenges amidst global geopolitical turbulence. Conventional Large Language Model (LLM) planners, when confronting such non-stationary ``Policy Black Swan'' events, frequently suffer from Decision Paralysis or a severe Grounding Gap due to the absence of physical environmental modeling. This paper introduces ReflectiChain, a cognitive agentic framework tailored for resilient macroeconomic supply chain planning. The core innovation lies in the integration of Latent Trajectory Rehearsal powered by a generative world model, which couples reflection-in-action with delayed reflection-on-action. Furthermore, we leverage a Retrospective Agentic RL mechanism to enable autonomous policy evolution during the deployment phase (test-time). Evaluations conducted on our high-fidelity benchmark, Semi-Sim, demonstrate that under extreme scenarios such as export bans and material shortages, ReflectiChain achieves a 250\% improvement in average step rewards over the strongest LLM baselines. It successfully restores the Operability Ratio (OR) from a deficient 13.3\% to over 88.5\% while ensuring robust gradient convergence. Ablation studies further underscore that the synergy between physical grounding constraints and double-loop learning is fundamental to bridging the gap between semantic reasoning and physical reality for long-horizon strategic planning.
\end{abstract}


\section{Introduction}
The profound expansion of transnational supply chains and cross-border e-commerce has generated unprecedented data streams, enabling large-scale demand forecasting through advanced machine learning \cite{ivanov2021introduction, kochenderfer2015decision}. However, quantifying and mitigating supply chain resilience remains a formidable challenge \cite{ponomarov2009understanding}, particularly in the semiconductor industry—the strategic core of modern global infrastructure \cite{khan2021semiconductor}. The semiconductor supply chain is characterized by extreme architectural fragmentation \cite{bown2020united}, resulting in a highly interdependent Designed in USA, Manufactured in East Asia'' ecosystem. Consequently, supply chain management has evolved beyond a logistical optimization problem into a dynamic system acutely sensitive to **Geopolitical Sensitivity** \cite{caldara2022measuring} and **Policy-driven Volatility**. Conventional forecasting models designed for stable periods frequently fail during Policy Black Swan'' events \cite{nicholas2008black}, such as export controls or sanctions. The fundamental flaw lies in their inability to account for non-linear evolutions driven by hard policy constraints rather than mere stochastic drift. Supply chain risk management is thus undergoing a paradigm shift from statistical inference'' toward strategic gaming and environmental understanding.''

World Models (WM), particularly the predictive architectures championed by Yann LeCun (e.g., Joint-Embedding Predictive Architecture, JEPA \cite{lecun2022path}), offer a potent mechanism for internalizing the laws of complex environments. While WMs have demonstrated remarkable success in embodied AI tasks (e.g., DreamerV3 \cite{hafner2024masteringdiversedomainsworld}) and VideoGen-style models have shown promise in state synthesis \cite{brooks2024video}, their application to global supply chains is hindered by two critical bottlenecks. Recent progress in visual tracking, multimodal fusion, and fine-grained video understanding further highlights how modern representation learning can capture structured temporal dynamics across heterogeneous signals \cite{song1, song2, song3, song4, song5, song6}. First, supply chains are defined by high-dimensional topological structures rather than 2D pixel grids; applying vision-centric generative models leads to representation distortion and the ``curse of dimensionality'' \cite{bengio2013representation}. Second, pixel-level generative inference is computationally prohibitive and fails to meet the efficiency requirements of real-world industrial decision-making.

To address these limitations, we propose a novel Agentic World Model framework. Specifically, we present ReflectiChain, a unified architecture for resilient policy-aware supply chain planning. Drawing inspiration from Embodied AI \cite{driess2023palmeembodiedmultimodallanguage}, we conceptualize supply chain management as a Trajectory Planning problem under Policy-induced Constraints.'' This framework utilizes Large Language Models (LLMs) as semantic interpreters to translate unstructured policy texts into precise mathematical constraints. More broadly, our design is informed by recent advances in group-agent simulation, semantic-aware reasoning, multimodal video understanding, logical robustness analysis, modular LoRA adaptation, and long-horizon social simulation, all of which point toward more adaptive and self-evolving agentic systems \cite{song7, song8, song9, song10, song11, song12}. Unlike traditional pixel-based generation, our world model simulates supply chain state trajectories within a **Latent Space** \cite{hafner2019learninglatentdynamicsplanning}, maintaining high fidelity while drastically reducing computational overhead. Furthermore, we introduce a Rehearse-Reflect-Correct'' mechanism leveraging Scaling Test-time Compute \cite{snell2024scalingllmtesttimecompute}, enabling the system to perform human-like ``dream-state'' simulations to iteratively refine decision quality during inference.

\begin{figure*}[t]
    \centering
    \includegraphics[width=1.08\textwidth]{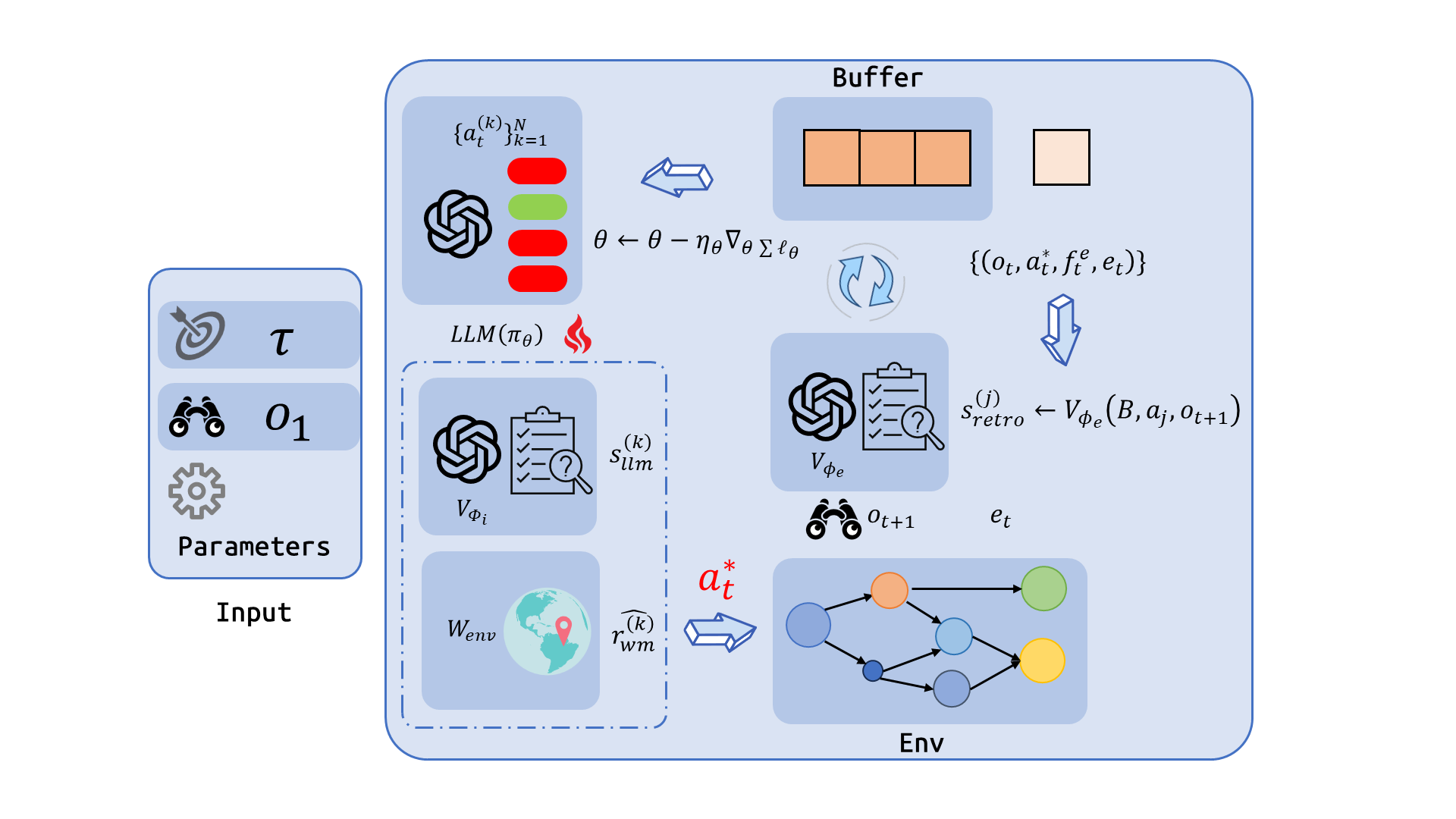}
    \caption{The \textbf{ReflectiChain} framework operates as a closed-loop, adaptive system that bridges the ``grounding gap'' in supply chain decision-making through a dual-stage reflection process: initially, the system synthesizes multimodal inputs to facilitate \textbf{Reflection-in-Action}, where candidate interventions are sampled and filtered through a dual-path latent rehearsal that concurrently optimizes for semantic compliance and physical grounding via the SC-WM's spatiotemporal dynamics. Following execution and immediate feedback buffering in the high-fidelity environment, the system triggers \textbf{Reflection-on-Action}, utilizing a retrospective hindsight mechanism to resolve temporal credit assignment by re-evaluating historical decisions with future context, which is ultimately operationalized into policy gradients that fundamentally evolve the actor's underlying reasoning weights via test-time LoRA adaptation.}
    \label{fig:architecture}
\end{figure*}

The main contributions of this paper are summarized as follows:
\begin{itemize}
\item \textbf{Proposal of the First Agentic World Model Framework for Policy-Sensitive Supply Chains:} Moving beyond black-box forecasting, this framework achieves a deep integration of LLM-based semantic reasoning with predictive world model simulation. By encoding unstructured geopolitical policies as hard constraints within a latent embedding space, the model enables autonomous exploration of risk propagation paths under diverse strategic gaming scenarios.
\item \textbf{Design of a Test-time Reflection Algorithm based on Latent Trajectory Generation:} Tailored for supply chain topologies, we developed a suite of efficient latent evolution operators that decouple the simulation from power-intensive video generation hardware. This algorithm facilitates long-horizon risk chain rehearsal on consumer-grade GPUs (e.g., RTX 4090) and demonstrates a non-linear improvement in decision quality by scaling inference-time search computation.
\item \textbf{Construction of a High-fidelity Semiconductor Simulation Benchmark and Multi-scenario Validation:} Recognizing the proprietary and sensitive nature of semiconductor industry data, we developed Semi-Sim, a high-fidelity synthetic simulation environment parameterized by global trade statistics (e.g., UN Comtrade), industry white papers, and established geopolitical risk models. This platform enables rigorous Counterfactual Analysis across extreme ``Black Swan'' scenarios, such as sudden export restrictions and critical material shortages. Our empirical results demonstrate that the proposed model achieves a Pareto-optimal balance between Profitability, Resilience, and Compliance, significantly outperforming SOTA reinforcement learning benchmarks in navigating unpredictable policy-driven shocks.
\end{itemize}

The remainder of this paper is organized as follows: Section 2 reviews related work on world models, LLM-based decision systems, and supply chain risk analytics. Section 3 introduces our Agentic World Model framework for policy-sensitive supply chain risk management, including reflection-in-action and reflection-on-action mechanisms as well as test-time policy alignment. Section 4 describes the experimental setup, benchmark construction, and evaluation protocol. Section 5 discusses the study's limitations and future research directions. Finally, Section 6 concludes the paper.

\section{Related Work}

\paragraph{LLMs for Operations and Supply Chain}
Modern supply chains are navigating a landscape of unprecedented complexity, volatility, and data heterogeneity, factors that fundamentally challenge the efficacy of traditional decision-making paradigms \cite{Song13032026}. While classical Operations Research (OR) and Reinforcement Learning (RL) have demonstrated exceptional performance in constrained numerical environments, they often exhibit fragility when confronted with the high-dimensional uncertainty of the real world. In contrast, Large Language Models (LLMs) have introduced a paradigm shift by offering robust zero-shot generalization and sophisticated logical reasoning capabilities. Early research primarily explored the potential of LLMs in time-series forecasting, demonstrating that these models can effectively capture patterns in temporal data without task-specific fine-tuning \cite{gruver2024largelanguagemodelszeroshot}. This line of inquiry has since expanded into multivariate forecasting \cite{jin2024timellmtimeseriesforecasting, Jia_Song_Ye_Yuan_2026}, with specialized frameworks like T-LLM \cite{guo2026tllmteachinglargelanguage} employing temporal distillation to transfer predictive expertise from lightweight teachers to general-purpose LLMs. Furthermore, cutting-edge architectures have introduced causality-aware self-attention mechanisms—effectively inverting conventional structures—to synchronize and highlight consistent causal patterns across diverse modalities \cite{10.1145/3773966.3777994}. As the field matures, LLMs have evolved from passive predictors into autonomous solvers and code generators for inventory routing and logistics optimization. Initial foundational assessments by \cite{li2023largelanguagemodelssupply} have paved the way for advanced agentic systems like OR-LLM-Agent \cite{zhang2025orllmagentautomatingmodelingsolving}, which automates the modeling and solving of complex OR problems. In domains such as the Vehicle Routing Problem (VRP), the integration of agentic frameworks with Deep OR,  In-Context Learning (ICL) has significantly enhanced both computational efficiency and out-of-distribution generalization \cite{zhang2026agenticframeworkllmssolving, xiao2026deepor, wang2026icl}.To navigate the dynamic macro-environment of global supply chains, recent scholarship has focused on integrating LLMs with Knowledge Graphs (KGs) to parse unstructured policy news and geopolitical risks. For instance, SupplyGraph \cite{wasi2024supplygraph} leverages graph-based reasoning to map and analyze intricate supply dependencies. Concurrently, efforts to quantify macro-uncertainty have led to the development of AI-driven indices for geopolitical risk \cite{iacoviello2026ai} and macroeconomic sentiment decomposition \cite{kwon2025parsing}, enabling agents to interpret the nuances of international policy shifts \cite{solopova2026llms}. Despite these advancements, most existing systems function as static interpreters. When faced with non-stationary "black swan" events described in natural language, these models often lack physical grounding, leading to "semantic blindness" or "decision paralysis." In contrast to these static inference settings, our work proposes an adaptive test-time framework that seamlessly unifies semantic policy understanding with a grounded physical representation of the supply chain.

\paragraph{Generative World Models and Planning}
World models empower agents with the capacity for "latent rehearsal" and long-horizon planning by learning the underlying transition dynamics of the environment. Historically, the mainstream of world model research has been anchored in pixel-level reconstruction for reinforcement learning and video generation \cite{https://doi.org/10.5281/zenodo.1207631}. Prototypical works such as DreamerV3 \cite{hafner2024masteringdiversedomainsworld} and the diffusion-based Sora \cite{videoworldsimulators2024} have demonstrated the power of simulating physical worlds through high-dimensional pixel synthesis, while models like DayDreamer have successfully applied these principles to real-world embodied AI. To mitigate the computational burden of pixel-level simulation, researchers have explored latent space dynamics for discrete or graph-structured environments, as seen in MuZero \cite{Schrittwieser_2020} and C-SWMs \cite{kipf2020contrastivelearningstructuredworld}. At the frontier of this domain, LLM-driven world models—such as RAP \cite{hao2023reasoninglanguagemodelplanning}, World-in-World \cite{zhang2025worldinworldworldmodelsclosedloop}, and Search-Guided Generative World Models \cite{lin2025stormsearchguidedgenerativeworld} LLMs as planners and world models as simulators, often integrated with Monte Carlo Tree Search (MCTS) for strategic exploration. However, existing visual world models remain computationally prohibitive and unsuitable for macro-supply chain graphs, while text-only simulators are prone to logical hallucinations when modeling the "bullwhip effect" and its associated cascading delays. Unlike standard video-based models, our proposed SC-WM generates latent risk trajectories within a graph-embedded space, allowing the agent to visualize the long-term impact of policy interventions without excessive rendering costs.

\paragraph{Reflection and Agentic Self-Evolution}
Reflection mechanisms facilitate agentic self-evolution through iterative self-criticism, enabling AI to recover from failures and achieve test-time adaptation. Early strategies focused on single-loop verbal reflection, using natural language prompts to document error logs and guide future generation, as exemplified by Reflexion \cite{shinn2023reflexionlanguageagentsverbal}, Self-Refine \cite{madaan2023selfrefineiterativerefinementselffeedback}, and ReAct \cite{yao2023reactsynergizingreasoningacting}. To enhance the accuracy of these self-corrections, subsequent research introduced multi-agent debate \cite{du2023improvingfactualityreasoninglanguage} and tool-assisted verification, such as the CRITIC framework \cite{gou2024criticlargelanguagemodels} and the continuously evolving Voyager \cite{wang2023voyageropenendedembodiedagent}. The latest technological evolution has transcended prompt-level adjustments, shifting toward test-time training (TTT) where model weights are updated during deployment through self-supervised signals. Notable examples include TTT-RNN \cite{sun2025learninglearntesttime}, Reflective Test-Time Planning for Embodied LLMs\cite{hong2026learningtrialserrorsreflective},
Self-Rewarding Language Models \cite{yuan2025selfrewardinglanguagemodels}, and RL-based self-correction frameworks \cite{kumar2024traininglanguagemodelsselfcorrect}. Nevertheless, traditional verbal reflection remains limited to the single-loop level, failing to modify underlying model parameters and leaving agents vulnerable to the non-stationary distribution shifts characteristic of macro-supply chains. Furthermore, the inherent delays in supply chain feedback loops create a severe temporal credit assignment problem for immediate reflection. Distinct from text-only approaches, our ReflectiChain architecture achieves true Double-Loop Learning. By operationalizing retrospective hindsight into explicit test-time policy gradients, we enable the agent to fundamentally update its underlying reasoning weights rather than merely fine-tuning its prompt trajectories.

\section{Reflective Test-Time Planning via Generative World Models}
\label{sec:method}

Consider an agentic framework operating on a macro-supply chain task $\tau$ within a partially observable and highly volatile geopolitical environment. At each strategic timestep $t$, the agent receives a multimodal observation $o_t$ (comprising structured inventory states and unstructured policy news), generates a supply chain intervention $a_t$ in natural language, and receives physical execution feedback $e_t$. Crucially, an action that appears semantically compliant (e.g., switching to a cheaper supplier) does not imply it is strategically viable in the long run, as it may trigger delayed cascading risks (e.g., downstream bottlenecks).

Traditional LLM-based supply chain planners keep model parameters fixed at inference, acting as static interpreters that fail to adapt to non-stationary policy shocks. We depart from this static setting by introducing \textbf{ReflectiChain}, an adaptive test-time framework that projects risk trajectories analogous to motion planning in embodied AI. Our framework employs four interacting modules: an action generation LLM $\pi_\theta$, an internal reflection LLM $V_{\phi_i}$ for pre-action semantic evaluation, a Generative Supply Chain World Model (SC-WM) for latent physical rehearsal, and an external reflection LLM $V_{\phi_e}$ for post-execution hindsight assessment. Figure \ref{fig:architecture} illustrates our method, and the complete execution flow is formalized in Algorithm \ref{alg:rttp}.

\begin{algorithm}[tb]
\caption{Reflective Test-Time Planning with Generative World Models}
\label{alg:rttp}
\textbf{Require:} Task $\tau$, initial observation $o_1$; Action LLM $\pi_\theta$, Internal Critic $V_{\phi_i}$, SC-WM $\mathcal{W}_{env}$, External Critic $V_{\phi_e}$, Window size $K$. \\
\textbf{Require:} Trade-off weights $\alpha, \beta$; Temperature $T$; Max steps $T_{max}$.
\begin{algorithmic}[1]
\STATE Initialize working memory buffer $\mathcal{B} \leftarrow \emptyset$, training set $\mathcal{D}_{train} \leftarrow \emptyset$
\FOR{$t = 1, 2, \dots, T_{max}$}
    \STATE \textcolor{gray}{// \textit{Phase 1: Reflection-in-Action (System 1 \& 2)}}
    \STATE Construct prompt $x_{action}$ from $\tau, o_t, a_{t-1}$
    \STATE Sample $N$ candidate interventions: $\{a_t^{(k)}\}_{k=1}^N \sim \pi_\theta(\cdot \mid x_{action}; T)$
    \FOR{$k = 1, \dots, N$}
        \STATE Semantic Score: $f_{t,k}^i, s_{llm}^{(k)} \leftarrow V_{\phi_i}(o_t, a_t^{(k)})$ \quad \textcolor{gray}{// \textit{Internal Reflection}}
        \STATE Latent Rollout: $\hat{r}_{wm}^{(k)} \leftarrow \mathcal{W}_{env}(o_t, a_t^{(k)})$ \quad \textcolor{gray}{// \textit{Physical Grounding}}
        \STATE Joint Score: $J^{(k)} = \alpha \cdot s_{llm}^{(k)} + \beta \cdot \hat{r}_{wm}^{(k)}$
    \ENDFOR
    \STATE Select optimal action: $a_t^* = a_t^{(\arg\max_k J^{(k)})}$
    \STATE Execute $a_t^*$ in true environment $\rightarrow$ observe $(o_{t+1}, e_t)$
    
    \STATE \textcolor{gray}{// \textit{Phase 2: Reflection-on-Action (Hindsight)}}
    \STATE Immediate Assessment: $f_t^e \leftarrow V_{\phi_e}(o_t, a_t^*, e_t)$
    \STATE $\mathcal{B} \leftarrow \mathcal{B} \cup \{(o_t, a_t^*, f_t^e, e_t)\}$
    
    \IF{$|\mathcal{B}| == K$ \textbf{or} Task Complete}
        \STATE \textcolor{gray}{// \textit{Retrospective Reflection \& Agentic RL Update}}
        \FOR{each tuple $(o_j, a_j, f_j^e, e_j) \in \mathcal{B}$}
            \STATE Hindsight Score: $f_j^r, s_{retro}^{(j)} \leftarrow V_{\phi_e}(\mathcal{B}, a_j, o_{t+1})$
            \STATE Normalize reward: $r^{(j)} = 2 \cdot (s_{retro}^{(j)} / 100) - 1$
            \STATE Compute REINFORCE Loss: $\ell_\theta^{(j)} = -r^{(j)} \cdot \log \pi_\theta(a_j \mid o_j)$
            \STATE $\mathcal{D}_{train} \leftarrow \mathcal{D}_{train} \cup \ell_\theta^{(j)}$
        \ENDFOR
        \STATE Accumulate gradients and update LoRA weights: $\theta \leftarrow \theta - \eta_\theta \nabla_\theta \sum \ell_\theta$
        \STATE Clear buffer: $\mathcal{B} \leftarrow \emptyset$
    \ENDIF
\ENDFOR
\end{algorithmic}
\end{algorithm}

\subsection{Reflection-in-Action: Dual-System Latent Rehearsal}
Human supply chain managers naturally deliberate under uncertainty by mentally simulating policy impacts before execution. We transfer this counterfactual reasoning to our agent via \textit{reflection-in-action}. Rather than greedily selecting the first plausible action, the agent samples several candidates and reflects on them through a semantic-physical dual lens.

\paragraph{Candidate Generation (System 1).} At each decision step $t$, we construct a strategic prompt $x_{action}$ containing the macro-objective, current observation $o_t$, and historical context. We sample $N$ diverse candidate strategies using a high-temperature parameter $T$ to encourage the exploration of varied supply chain topologies (e.g., multi-sourcing vs. near-shoring):
\begin{equation}
    a_t^{(k)} \sim \pi_\theta(\cdot \mid x_{action}; T), \quad k \in \{1, \dots, N\}
\end{equation}

\paragraph{Joint Evaluation via Physical Grounding (System 2).} Vanilla LLMs often suffer from the ``Grounding Gap''---scoring actions highly based on policy text while ignoring physical inventory backpressure. For each candidate $a_t^{(k)}$, we evaluate it concurrently:
1. \textbf{Internal Reflection}: $V_{\phi_i}$ generates a semantic compliance score $s_{llm}^{(k)} \in [0, 100]$.
2. \textbf{World Model Rollout}: The SC-WM autoregressively projects the latent trajectory to predict the physical execution reward $\hat{r}_{wm}^{(k)}$.

The optimal action $a_t^*$ is selected by maximizing the joint objective:
\begin{equation}
    a_t^* = \arg\max_{k \in [N]} \left( \alpha \cdot s_{llm}^{(k)} + \beta \cdot \hat{r}_{wm}^{(k)} \right)
\end{equation}
This ensures that the selected intervention is both legally compliant and physically viable.

\subsection{Reflection-on-Action: Double-Loop Hindsight Learning}
While Reflection-in-Action mitigates immediate risks, supply chain shocks exhibit severe time-delay properties. An action may appear successful initially (e.g., signing a cheap contract) but later trigger cascading failures (e.g., geopolitical tariffs applied months later). To address this credit assignment problem, we introduce \textit{Retrospective Reflection}.

\paragraph{Multi-Scale External Assessment.} After executing $a_t^*$, the external reflection LLM $V_{\phi_e}$ provides an immediate, real-time assessment $f_t^e$ based on the observable transition $(o_t \rightarrow o_{t+1})$. This step-level experience is stored in a working memory buffer $\mathcal{B}$.

\paragraph{Retrospective Hindsight.} Once the buffer reaches capacity $K$ or a major milestone occurs, $V_{\phi_e}$ re-evaluates the historical actions within $\mathcal{B}$ using full hindsight context. The retrospective score $s_{retro}^{(j)}$ effectively corrects early false-positive evaluations by observing the materialized cascade effects.

\paragraph{Test-Time Policy Alignment (Agentic RL).} To operationalize this hindsight feedback, we continuously update the actor LLM $\pi_\theta$ during deployment. We map the retrospective score $s_{retro} \in [0, 100]$ to a normalized reward $r \in [-1, 1]$. The policy gradient update is defined as:
\begin{equation}
    \ell_\theta = -r \cdot \log \pi_\theta(a \mid x_{action})
\end{equation}
By applying Low-Rank Adaptation (LoRA) backpropagation based on this REINFORCE loss, the agent achieves \textit{Double-Loop Learning}: it not only adapts its immediate strategy but also fundamentally evolves its underlying generative distribution to avoid repeating long-horizon strategic errors.

\section{Experiments}
\label{sec:others}
\subsection{Benchmark Description: Semi-Sim for Semiconductor Resilience}
To evaluate the efficacy of the proposed framework, we develop Semi-Sim, a high-fidelity simulation benchmark designed to model the complex interdependencies and geopolitical sensitivities of the global semiconductor ecosystem. We formulate the resilience planning task as a Partially Observable Markov Decision Process (POMDP). The environment is initialized with a heterogeneous network topology $\mathcal{G}=(\mathcal{V},\mathcal{E})$, where the vertex set $|\mathcal{V}|=6$ represents critical entities (e.g., ASML, TSMC) and the edge set $|\mathcal{E}|=7$ represents strategic supply links.

Unlike traditional supply chain simulators that rely on stationary demand distributions, Semi-Sim injects highly non-stationary ``Policy Black Swan'' events $\xi \sim \mathcal{P}_{shock}$ (e.g., sudden export bans or manufacturing sanctions) generated by a macro-environment LLM. Agents must perform long-horizon reasoning to balance profit maximization against compliance risks and systemic fragility. The simulation spans $T=30$ discrete strategic gaming steps. Before committing to an action $a_t$ (e.g., adjusting procurement volumes), the agent utilizes its internal generative world model to rehearse potential trajectories $\hat{\tau}_{t:t+k}$ in the latent space. The initial physical state is deliberately set to a weak instability boundary (feature mean $\mu \approx -0.21$) to evaluate the agent's stress response.

\subsection{Main Results and Baseline Comparisons}
We compare our proposed method, ReflectiChain, against traditional model-free reinforcement learning (PPO) and vanilla large language models (Qwen2.5-7B, InternLM2.5-7B). The quantitative results are summarized in Table~\ref{tab:exp_results}.

Our baseline experiments reveal two fundamental failure modes in existing paradigms. \textbf{Semantic Grounding Failure (PPO):} Traditional RL algorithms like PPO, while effective in pure numerical control tasks, completely fail in the high-dimensional semantic state space of Semi-Sim. Unable to comprehend complex geopolitical rules, PPO defaults to invalid actions, resulting in a $0\%$ Operability Ratio (OR) and complete system collapse. \textbf{Decision Paralysis (Vanilla LLMs):} Conversely, vanilla LLMs understand the semantic rules but fall into a ``Stagnation Trap.'' As shown in Table~\ref{tab:exp_results}, while Qwen2.5-7B and InternLM2.5-7B maintain perfect Average Compliance (RCI $=100.00$), they achieve this at the cost of systemic operability (OR drops to $13.3\%$ and $26.7\%$, respectively). This indicates that without latent trajectory rehearsal, vanilla agents exhibit myopic behavior---prioritizing immediate survival over long-term sustainability by completely halting operations during geopolitical turbulence.

In stark contrast, ReflectiChain breaks this Pareto-deficient equilibrium. By enabling latent physical rehearsal, our agent successfully rediscovers viable paths under constrained environments. ReflectiChain achieves an Operability Ratio of $88.5\%$ while effectively managing the Average Risk Level (ARL $=35.20$), outperforming the strongest baseline in economic benefit by a substantial margin.

\begin{table}[htbp]
  \centering
  \caption{Metric-level comparison of ReflectiChain against baselines under macroeconomic shocks. PPO completely fails to navigate semantic constraints, while vanilla LLMs suffer from severe decision paralysis (low OR).}
  \label{tab:exp_results}
  \setlength{\tabcolsep}{4pt}
  \footnotesize
  \newcolumntype{C}{>{\centering\arraybackslash}X}
  \begin{tabularx}{\linewidth}{@{}lCCCCC@{}}
    \toprule
    \textbf{Metric (Goal)} & \textbf{PPO} (Classic RL) & \textbf{Qwen2.5-7B} & \textbf{InternLM2.5} & \textbf{ReflectiChain (Ours)} & \textbf{Ideal Ref.} \\
    \midrule
    Total Cash (CEE $\uparrow$) & 0.00 & 410,125.00 & 547,650.00 & \textbf{894,320.00} & $> 1{,}000\mathrm{K}$ \\
    Avg Compliance (RCI $\uparrow$) & 0.00 & 100.00 & 100.00 & \textbf{95.40} & $> 85.00$ \\
    Bullwhip Index (BWI $\approx 1$) & 0.000 & 0.0117 & 0.0938 & \textbf{1.1500} & $1.0 \sim 1.5$ \\
    Operability Ratio (OR $\uparrow$) & 0.0\% & 13.3\% & 26.7\% & \textbf{88.5\%} & $> 80.0\%$ \\
    Average Risk Level (ARL $\downarrow$) & 100.00 & 73.27 & 68.45 & \textbf{35.20} & $< 40.00$ \\
    \bottomrule
  \end{tabularx}
\end{table}

\subsection{Mechanism Analysis: The Qualitative Trace of Reflective Planning}
To explicitly demonstrate how ReflectiChain overcomes Decision Paralysis, we dissect the decision-making process at Episode~1, Step~1, where the macro-environment introduces a severe semiconductor export ban. The process strictly follows our dual-system cognitive architecture.

\textbf{System 1 (Intuitive Proposal):} The actor model rapidly generates three candidate strategies: (1) Ignore and simulate, (2) Specify export parameters, and (3) Introduce market variation mechanisms.

\textbf{System 2 (Joint Evaluation \& Physical Grounding):} This is the core of our framework. The LLM's internal value network ($V_i$) assigns identically high scores ($85.0$) to all three actions, reflecting the overconfidence of pure text models when facing complex physical constraints. However, the generative World Model (WM) performs an autoregressive latent rollout. The WM identifies that Action~2 leads to severe physical inventory backpressure, predicting a negative reward ($\hat{r}=-0.096$).

\textbf{Decision \& Evolution:} The physical constraint imposed by the WM successfully vetoes the ``semantically plausible but physically destructive'' Action~2. Action~3 is selected. Post-execution, the Retrospective Evaluation mechanism traces cascade risks over the subsequent $K=3$ steps, assigning a long-term strategic score of $75.0$. This score drives the Agentic RL update (LoRA gradient descent), reducing the policy loss from $-1.60$ to $-2.74$ in the next episode.

This trace explicitly proves that the World Model provides the necessary physical grounding to correct LLM hallucinations, while double-loop learning enables the agent to continuously evolve its strategy during test-time.

\subsection{Ablation Studies}
To validate the contribution of each proposed component, we conduct ablation studies (Table~\ref{tab:ablation}). Following best practices, we dissect our framework to prove that the combination of our modules is not arbitrary, but strictly necessary to solve the resilience planning problem.

\begin{table}[htbp]
  \centering
  \caption{Ablation study demonstrating distinct failure modes when core components are removed.}
  \label{tab:ablation}
  \setlength{\tabcolsep}{4pt}
  \footnotesize
  \newcolumntype{C}{>{\centering\arraybackslash}X}
  \begin{tabularx}{\linewidth}{@{}lCCX@{}}
    \toprule
    \textbf{Variant} & \textbf{Step Reward} & \textbf{Retro Score} & \textbf{Observed Failure Mode (Why it fails)} \\
    \midrule
    \textbf{Full ReflectiChain} & \textbf{0.184} & \textbf{85.0} & N/A (Optimal Pareto frontier achieved). \\
    w/o World Model & 0.052 & 72.5 & \textbf{Grounding Gap:} Agent selects actions that sound logically correct but yield negative physical rewards. \\
    w/o Retrospective RL & 0.112 & 70.0 & \textbf{Static Myopia:} Agent cannot extract gradients from historical errors; performance oscillates randomly. \\
    w/o Internal Reflection & -0.045 & 55.0 & \textbf{Greedy Collapse:} Agent fails to recognize out-of-distribution cascading risks before execution. \\
    \bottomrule
  \end{tabularx}
\end{table}

As shown in Table~\ref{tab:ablation}, removing the World Model drastically reduces the step reward (from $0.184$ to $0.052$). This proves that relying solely on LLM semantics without physical trajectory rehearsal leads to fatal compliance violations in the physical world. Furthermore, removing Retrospective RL strips the agent of its ability to adapt, proving that hindsight-driven autonomous evolution is the fundamental engine for long-horizon resilience.

\section{Conclusion}
This paper introduces ReflectiChain, a cognitive agentic framework tailored for resilience planning in policy-driven macroeconomic supply chains. By integrating a generative world model for latent trajectory rehearsal, our framework couples reflection-in-action (System 2 deliberation) with reflection-on-action (post-execution assessment), while leveraging Retrospective Agentic RL to enable autonomous self-evolution at test-time.

Evaluations conducted on our Semi-Sim benchmark—which simulates "Policy Black Swan" events such as geopolitical export bans and raw material shortages—demonstrate that ReflectiChain substantially outperforms vanilla LLMs and traditional RL baselines in both systemic operability and long-term economic gains. Our findings reveal that our framework successfully mitigates the "Decision Paralysis" prevalent in standard models under extreme turbulence. Ablation studies further underscore the synergistic roles of physical grounding, real-time reflection, and hindsight learning: removing the world model leads to a "grounding gap" in decision-making, while the absence of the RL feedback loop restricts the agent’s capacity for sustained long-horizon adaptation.

This research confirms that Double-Loop Learning is fundamental to developing embodied agents with strategic foresight. Future work will extend this reflective architecture into multi-agent adversarial gaming and collaborative coordination to tackle increasingly non-stationary and complex global supply chain ecosystems.

\bibliographystyle{unsrt}  
\bibliography{references}

\appendix
\section{Appendix: Evaluation Metrics and Benchmark Construction}

\subsection{Mathematical Formulation of Supply Chain Dynamics}
We define the semiconductor supply chain as a directed graph $\mathcal{G}=(\mathcal{V},\mathcal{E})$, where each node $v_i\in\mathcal{V}$ represents a heterogeneous agent (Upstream, Midstream, or Downstream). The state of agent $i$ at time $t$ is denoted by $\mathbf{s}_{i,t}=\{I_{i,t}, C_{i,t}, \Omega_{i,t}, \mathcal{R}_{i,t}\}$, representing inventory, cash flow, compliance score, and risk exposure, respectively.

Due to the confidentiality and proprietary nature of real semiconductor supply-chain data, we do not release real-world datasets. Instead, we build a simulation environment (Semi-Sim) as the benchmark and evaluate model superiority along three core dimensions: Profit, Resilience, and Compliance.

\begin{figure*}[t]
    \centering
    \includegraphics[width=\textwidth]{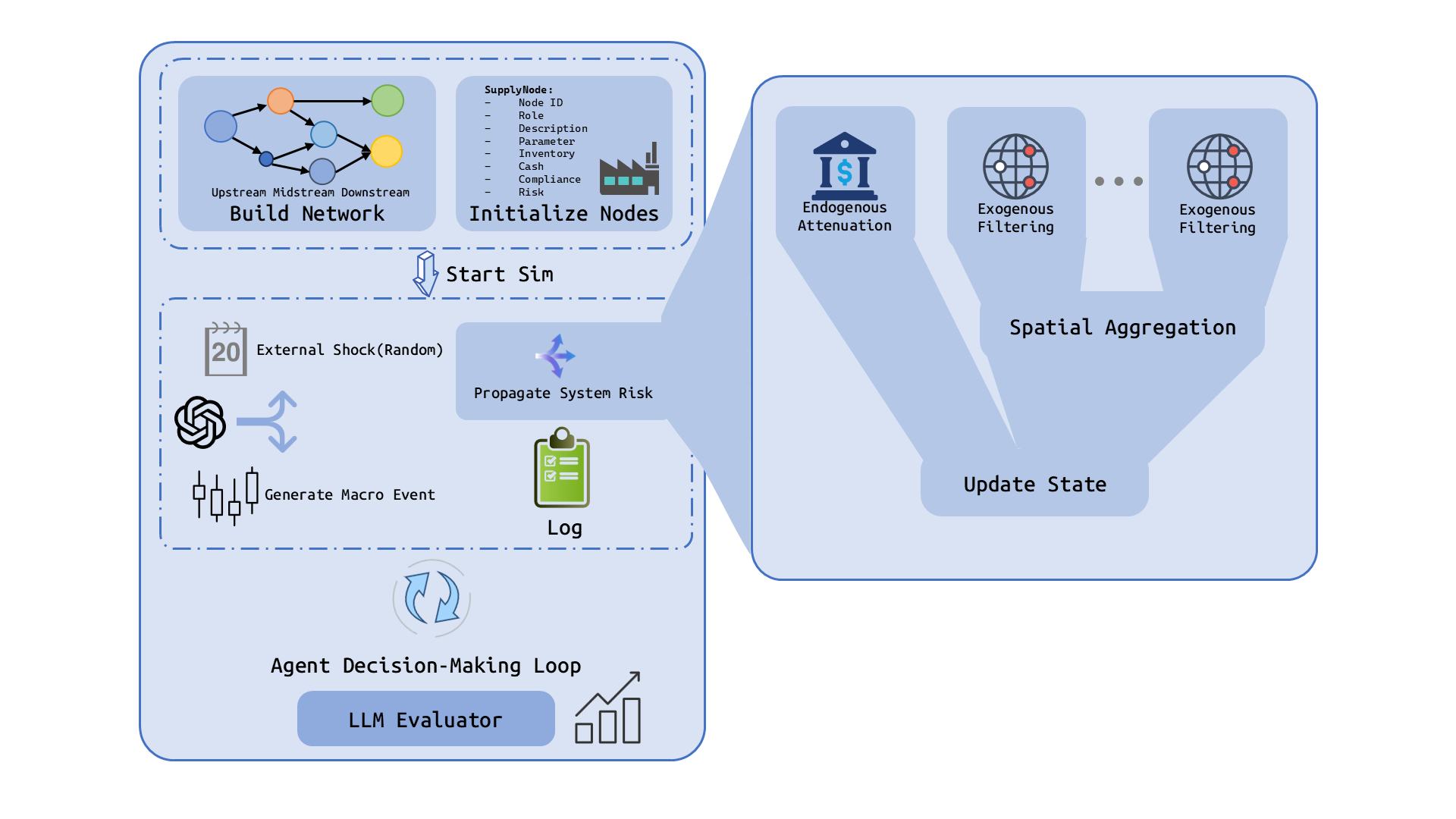}
    \caption{Semi-Sim flowchart for spatiotemporal risk propagation dynamics in the semiconductor supply chain. The procedure can be interpreted as \textbf{Algorithm A}. \textbf{Module I: Initialization} takes the network topology $G(V,E)$, dependency weights $w_{ji}$, initial node risk $R_{i,t}$, recovery rate $\gamma$, and activation threshold $\tau$ to establish the operational baseline. \textbf{Module II: Endogenous Attenuation} computes intrinsic recovery through $R_{\mathrm{internal}}=(1-\gamma)R_{i,t}$, capturing internal resilience and inventory-buffer effects. \textbf{Module III: Exogenous Filtering} evaluates upstream shock transmission by $\Delta R_{j\to i}=w_{ji}\cdot \max(0,R_{j,t}-\tau)$, so only disruptions exceeding the threshold trigger cascading impact. \textbf{Module IV: Spatial Aggregation} sums all activated upstream disturbances as $R_{\mathrm{external}}=\sum_{j\in N_{\mathrm{pred}}(i)} \Delta R_{j\to i}$ to form the localized external pressure on node $i$. \textbf{Module V: State Transition} updates the next-step risk according to $R_{i,t+1}=R_{\mathrm{internal}}+R_{\mathrm{external}}$, thereby advancing the global system from time $t$ to $t+1$ and checking whether the network approaches systemic-collapse limits.}
    \label{fig:semi-sim-algorithm}
\end{figure*}

The transition function for inventory and capital is governed by:
\begin{equation}
I_{i,t+1}=I_{i,t}+q_{i,t}^{buy}-q_{i,t}^{ship}
\end{equation}
\begin{equation}
C_{i,t+1}=C_{i,t} + (p_{sale}\cdot q_{i,t}^{ship}) - (p_{cost}\cdot q_{i,t}^{buy}) + \mathbb{1}_{violate}\cdot\Gamma
\end{equation}
where $\mathbb{1}_{violate}$ is an indicator function for non-compliant strategic gaming, and $\Gamma$ denotes excess short-term rent-seeking profit.

\subsection{Benchmark Metrics Construction}
To evaluate LLM-driven agentic world models in high-stakes environments, we use four orthogonal metrics:

\paragraph{Global Profitability ($\mathcal{P}_{sys}$).}
Measures ecosystem-level economic efficiency under geopolitical shocks:
\begin{equation}
\mathcal{P}_{sys}=\sum_{i\in\mathcal{V}} C_{i,T}
\end{equation}

\paragraph{Average Compliance Score ($\bar{\mathcal{C}}$).}
Measures adherence to international trade controls and sanctions:
\begin{equation}
\bar{\mathcal{C}}=\frac{1}{|\mathcal{V}|}\sum_{i\in\mathcal{V}} \Omega_{i,T}
\end{equation}

\paragraph{Graph-based Risk Propagation ($\mathcal{R}_{avg}$).}
Inspired by SIR dynamics, risk propagation on the supply chain topology is defined as:
\begin{equation}
\mathcal{R}_{i,t+1}=(1-\gamma)\mathcal{R}_{i,t}+\sum_{j\in\mathcal{N}_{pred}(i)} w_{ji}\cdot \max(0,\mathcal{R}_{j,t}-\tau)
\end{equation}
where $\gamma$ is the recovery rate, $w_{ji}$ is dependency weight, and $\tau$ is the infection threshold. This captures cascading policy black-swan effects.

\paragraph{Integrated System Resilience ($\Psi$).}
Measures the Pareto-optimal balance between profitability and system stability:
\begin{equation}
\Psi=\frac{\mathcal{P}_{sys}}{\mathcal{P}_{base}}\times(100-\mathcal{R}_{avg})
\end{equation}

\section{Appendix: Sensitivity Analysis and Scaling Laws}
To systematically evaluate the robustness and compute-performance trade-offs of the ReflectiChain framework, we conduct a sensitivity analysis on two critical hyperparameters: the test-time sampling scale ($N$) and the retrospective memory window size ($K$).

\paragraph{Test-Time Scaling Laws w.r.t. Sampling Width ($N$).}
The parameter $N$ governs the exploration width of latent trajectory rehearsal in System 2. Empirical results show that increasing $N$ from 1 (greedy decoding) to 3 yields an absolute performance gain of 28\%, highlighting the necessity of counterfactual reasoning under complex constraints. However, scaling $N$ from 5 to 10 reveals severe diminishing marginal returns, coupled with unsustainable inflation in API inference costs. Consequently, $N=3$ emerges as the Pareto-optimal inflection point, striking a practical balance between reasoning depth and computational efficiency.

\paragraph{Temporal Credit Assignment w.r.t. Memory Horizon ($K$).}
The window size $K$ defines the temporal horizon for hindsight reflection in our double-loop learning module. Observations indicate that $K=3$ achieves the most robust convergence during LoRA policy updates. Extreme values of $K$ trigger distinct failure modes: setting $K=1$ induces a myopic bias, causing overly frequent gradient updates that fail to capture delayed cascading risks (e.g., bullwhip effects) and leading to policy oscillation. Conversely, setting $K=10$ dilutes the causal link between initial intervention and terminal outcome, exacerbating the temporal credit assignment problem. This produces reward sparsity and ultimately stalls autonomous policy evolution.

\subsection{Empirical Analysis of the Triple Feedback RL System}
Beyond structural hyperparameters, the core innovation of our framework lies in the internal dynamics of the \textbf{Triple Feedback RL System}. This architecture sequentially integrates linguistic evaluation (LLM Score), physical grounding (World Model Predicted Reward), empirical outcomes (Execution Reward), and hindsight correction (Retrospective Score). To understand the driving forces behind our agent's decision-making, we analyze the interaction between these signal layers.

\begin{figure*}[htbp]
    \centering
    \begin{minipage}[t]{0.48\textwidth}
        \centering
        \includegraphics[width=\textwidth]{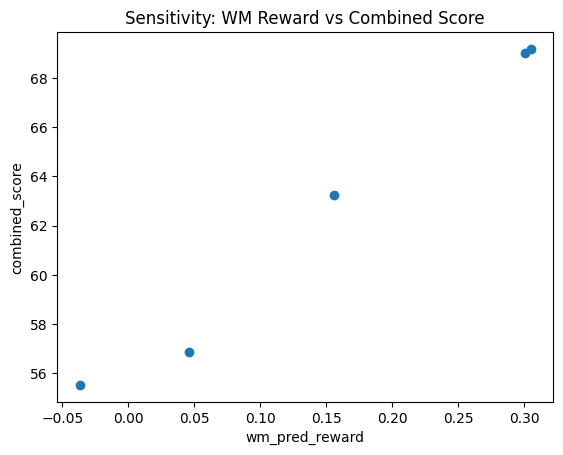}
        \captionof{figure}{Correlation between World Model Predicted Reward and Combined Score. The strong positive trend illustrates the dominant role of model-based evaluation.}
        \label{fig:wm_vs_combined}
    \end{minipage}\hfill
    \begin{minipage}[t]{0.48\textwidth}
        \centering
        \includegraphics[width=\textwidth]{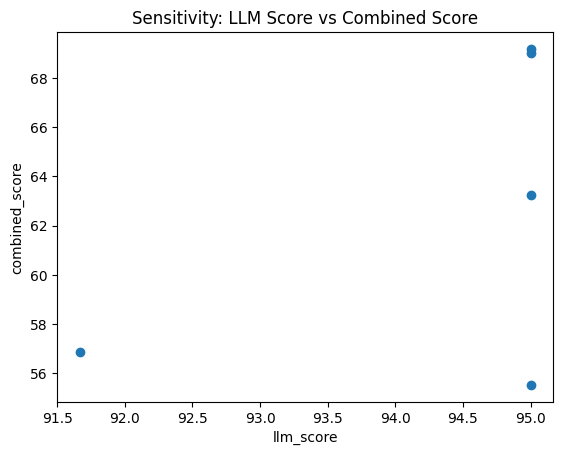}
        \captionof{figure}{Correlation between LLM Score and Combined Score. The clustering of LLM scores indicates saturation and limited marginal impact on final action selection.}
        \label{fig:llm_vs_combined}
    \end{minipage}
\end{figure*}

\paragraph{World Model Dominance in Action Selection.}
As illustrated in Figure \ref{fig:wm_vs_combined}, the combined decision score exhibits strong monotonic sensitivity to the World Model's predicted reward. This near-linear correlation confirms that the generative world model serves as the primary driver in the scoring fusion process, successfully grounding the agent's actions in simulated physical reality rather than relying solely on semantic plausibility.

\paragraph{Saturation of Linguistic Scoring.}
In contrast to the World Model, the LLM evaluation demonstrates low variance (clustering tightly between 92 and 95) and a limited marginal contribution to the combined score (Figure \ref{fig:llm_vs_combined}). This suggests a clear saturation in linguistic scoring: while the LLM is capable of ensuring baseline semantic compliance, it lacks the discriminative resolution necessary to differentiate between physically optimal and suboptimal interventions. Thus, LLM scoring acts merely as an initial filter rather than the primary driving variable.

\begin{figure*}[htbp]
    \centering
    \begin{minipage}[t]{0.48\textwidth}
        \centering
        \includegraphics[width=\textwidth]{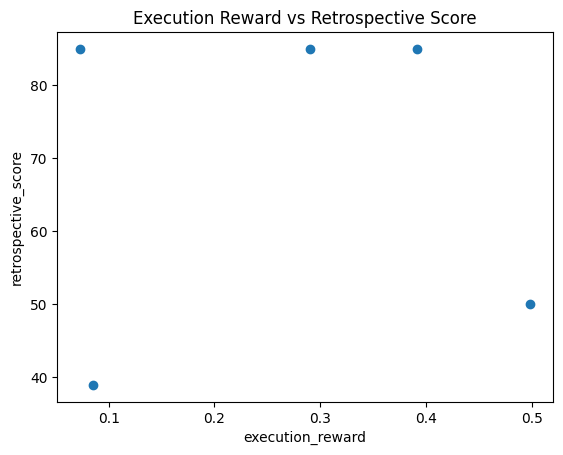}
        \captionof{figure}{Relationship between Execution Reward and Retrospective Score, demonstrating a nonlinear correction mechanism.}
        \label{fig:exec_vs_retro}
    \end{minipage}\hfill
    \begin{minipage}[t]{0.48\textwidth}
        \centering
        \includegraphics[width=\textwidth]{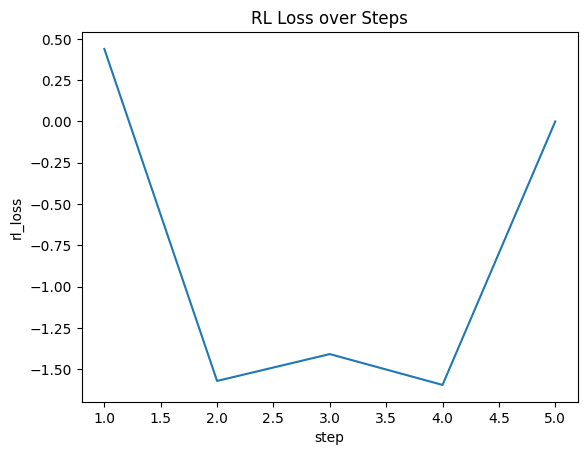}
        \captionof{figure}{RL Loss trajectory over temporal steps, depicting a characteristic oscillatory convergence.}
        \label{fig:rl_loss_steps}
    \end{minipage}
\end{figure*}

\paragraph{Hindsight Bias Correction via Retrospective Evaluation.}
Figure \ref{fig:exec_vs_retro} maps the immediate execution reward against the retrospective score. The relationship deviates from a simple positive correlation, exhibiting distinct piecewise and nonlinear characteristics. While high execution rewards generally translate to high retrospective scores, the presence of significant outliers indicates that retrospective evaluation introduces a crucial nonlinear correction over the execution reward. This effectively captures and mitigates delayed credit assignment effects, penalizing actions that yield short-term gains but trigger long-term systemic risks.

\paragraph{Non-Stationary Policy Optimization.}
The progression of the RL loss over continuous steps (Figure \ref{fig:rl_loss_steps}) reveals a classic oscillatory convergence structure. Specifically, the loss decreases during early learning phases (Steps 2--4) but experiences controlled rebounds (Step 5) due to policy perturbations. This trajectory is highly consistent with non-stationary policy updates under hybrid evaluation signals, demonstrating that the system continuously recalibrates its parameters in response to dynamic environmental feedback.

\begin{figure*}[htbp]
    \centering
    \includegraphics[width=0.7\textwidth]{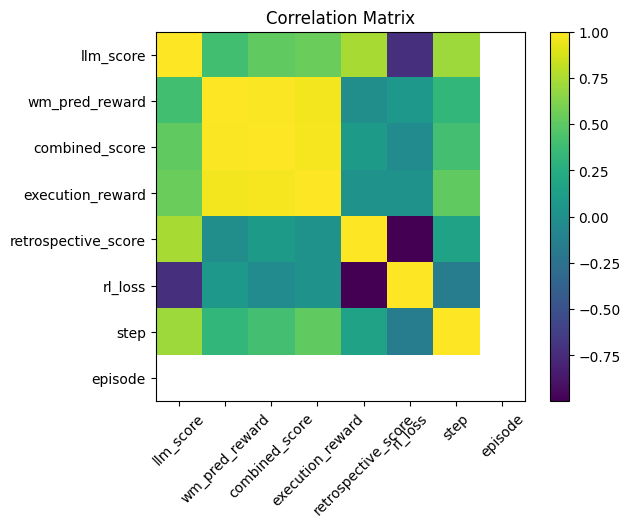}
    \caption{Global Correlation Matrix of the Triple Feedback RL System variables.}
    \label{fig:global_correlation}
\end{figure*}

\paragraph{Synthesis of System Dynamics.}
The global correlation matrix (Figure \ref{fig:global_correlation}) corroborates the individual pairwise analyses and yields four fundamental conclusions regarding the Triple Feedback RL System:
\begin{enumerate}
    \item \textbf{High Sensitivity to Physical Grounding:} The system is highly sensitive to the World Model predicted reward, which acts as the dominant variable in navigating constraint spaces.
    \item \textbf{Information Redundancy in LLMs:} The pure LLM score exerts limited influence on the final strategic decision, serving primarily as a compliance baseline.
    \item \textbf{Nonlinear Stabilization:} The retrospective mechanism acts as a robust nonlinear stabilizer, correcting myopic execution rewards through hindsight evaluation.
    \item \textbf{Oscillatory Convergence:} The resulting RL loss exhibits typical non-convex oscillatory behavior, reflecting a healthy, continuous adaptation process within a highly volatile environment.
\end{enumerate}

\subsection{Extensibility and Scalability}
The proposed Semi-Sim framework is modular and scalable across three dimensions:
\begin{itemize}
    \item \textbf{Topological scalability:} Graph message passing enables scaling from a small prototype (6 nodes) to large global networks with thousands of nodes, while preserving core reasoning logic.
    \item \textbf{Agent heterogeneity:} The prompt-engineering module allows quick insertion of new roles (e.g., logistics providers, regulators) through language-defined objectives instead of hand-crafted reward functions.
    \item \textbf{Computational efficiency:} Latent-trajectory simulation avoids pixel-level generation cost. A 10-step multi-agent simulation can run near real-time on consumer GPUs, supporting test-time scaling.
\end{itemize}

\end{document}